\pdfoutput=1
\label{key}
\documentclass[letterpaper, 10 pt, conference]{ieeeconf}  

\IEEEoverridecommandlockouts                              

\usepackage{graphicx} 
\usepackage{epsfig} 
\usepackage{epstopdf}
\usepackage{mathptmx} 
\usepackage{amsmath} 
\usepackage{amssymb}  
\usepackage{hyperref}

\newcommand{\vect}[1]{\boldsymbol{#1}}
\newcommand{\bR}{\mathbf{R}}
\newcommand{\bE}{\mathbb{E}}
\newcommand{\Lapl}{\mathcal{L}}

\newcommand{\beginsupplement}{%
	\setcounter{table}{0}
	\setcounter{section}{0}
	\renewcommand{\thetable}{S\arabic{table}}%
	\setcounter{figure}{0}
	\renewcommand{\thefigure}{S\arabic{figure}}%
}

\usepackage [english]{babel}
\usepackage [autostyle, english = american]{csquotes}
\MakeOuterQuote{"}

\usepackage{algpseudocode}
\usepackage{algorithm}

\title{\LARGE \bf
Fast, Robust, Continuous Monocular Egomotion Computation
}

\author{Andrew Jaegle$^*$, Stephen Phillips$^*$, and Kostas Daniilidis\\
\thanks{$^*$Authors contributed equally.}
University of Pennsylvania\\
Philadelphia, PA USA \\ 
{\tt\small \{ajaegle, stephi, kostas\}@seas.upenn.edu}
}

\begin{document}

\maketitle
\thispagestyle{empty}
\pagestyle{empty}

\begin{abstract}

We propose robust methods for estimating camera egomotion in noisy, real-world monocular image sequences in the general case of unknown observer rotation and translation with two views and a small baseline. This is a difficult problem because of the nonconvex cost function of the perspective camera motion equation and because of non-Gaussian noise arising from noisy optical flow estimates and scene non-rigidity. To address this problem, we introduce the expected residual likelihood method (ERL), which estimates confidence weights for noisy optical flow data using likelihood distributions of the residuals of the flow field under a range of counterfactual model parameters. We show that ERL is effective at identifying outliers and recovering appropriate confidence weights in many settings. We compare ERL to a novel formulation of the perspective camera motion equation using a lifted kernel, a recently proposed optimization framework for joint parameter and confidence weight estimation with good empirical properties. We incorporate these strategies into a motion estimation pipeline that avoids falling into local minima. We find that ERL outperforms the lifted kernel method and baseline monocular egomotion estimation strategies on the challenging KITTI dataset, while adding almost no runtime cost over baseline egomotion methods.

\end{abstract}

\section{Introduction}


\begin{figure}[tpb]
	\centering
	\includegraphics[scale=0.25]{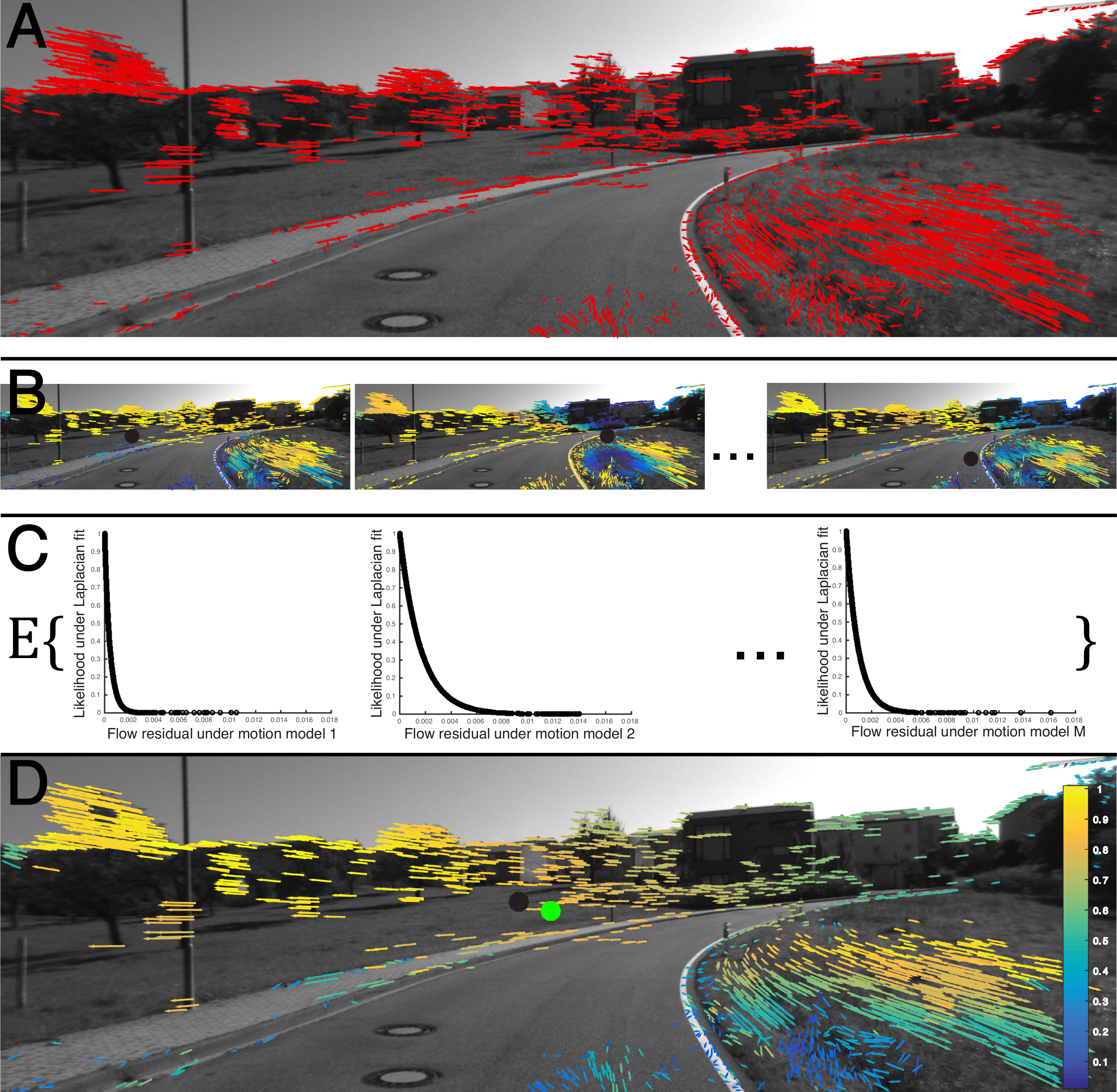}
	\caption{Schematic depiction of the ERL method for egomotion estimation from noisy flow fields. Figure best viewed in color. (A) Example optical flow field from two frames of KITTI odometry (sequence 5, images 2358-2359). Note the outliers on the grass in the lower right part of the image and scattered throughout the flow field. (B) We evaluate the flow field under $M$ models with translation parameters sampled uniformly over the unit hemisphere. The residuals for the flow field under three counterfactual models are shown. Each black point indicates the translation direction used. Residuals are scaled to [0,1] for visualization. (C) We estimate the likelihood of each observed residual under each of the models by fitting a Laplacian distribution to each set of residuals. The final confidence weight for each flow vector is estimated as the expected value of the residual likelihood over the set of counterfactual models. Likelihood distributions are shown for the three models above. (D) The weighted flow field is used to make a final estimate of the true egomotion parameters. The black point indicates the translation direction estimated using ERL and the green point indicates ground truth. The unweighted estimate of translation is not visible as it is outside of the image bounds.}
	\label{figurelabel}
\end{figure}

Visual odometry in real-world situations has attracted increased attention in the past few years in large part because of its applications in robotics domains such as autonomous driving and unmanned aerial vehicle (UAV) navigation. Stereo odometry and simultaneous localization and mapping (SLAM) methods using recently introduced depth sensors have made dramatic progress on real-world datasets. Significant advances have also been achieved in the case of \textit{monocular} visual odometry when combined with inertial information. 

State-of-the-art visual odometry uses either the discrete epipolar constraint to validate feature correspondences and compute inter-frame motion \cite{scaramuzza_visual_2011} or directly estimates 3D motion and 3D map alignment from image intensities \cite{forster_svo:_2014}. In contrast to the state of the art, in this paper we revisit the \textit{continuous} formulation of structure from motion (SfM), which computes the translational and rotational velocities and depths up to a scale from optical flow measurements. Our motivation lies in several observations: 
\begin{itemize}
	\item
	UAV control schemes often need to estimate the translational velocity, which is frequently done using a combination of monocular egomotion computations and single-point depths from sonar \cite{bristeau2011navigation}.
	\item
	Fast UAV maneuvers require an immediate estimate of the direction of translation (the focus of expansion) in order to compute a time-to-collision map.
	\item
	Continuous SfM computations result in better estimates when the incoming frame rate is high and the baseline is very small.
	\end{itemize}

However, estimating camera motion and scene parameters from a single camera (\textit{monocular} egomotion estimation) remains a challenging problem. This problem case arises in many contexts where sensor weight and cost are at a premium, as is the case for lightweight UAVs and consumer cameras. Situations involving monocular sensors on small platforms pose additional problems: computational resources are often very limited and estimates must be made in real time under unusual viewing conditions (e.g. with a vertically flipped camera, no visible ground plane, and a single pass through a scene). These contexts present many sources of noise. Real-time flow estimation produces unreliable data, and the associated noise is often pervasive and non-Gaussian, which makes estimation difficult and explicit outlier rejection problematic. Furthermore, violations of the assumption of scene rigidity due to independent motion of objects in the scene can lead to valid flow estimates that are outliers nonetheless. Even in the noise-free case, camera motion estimation is plagued with many suboptimal interpretations (illusions) caused by the hilly structure of the cost function. Additionally, forward motion, which is very common in real-world navigation, is known to be particularly hard for monocular visual odometry \cite{oliensis_least-squares_2005}. 


We propose an algorithm suitable for the robust estimation of camera egomotion and scene depth from noisy flow in real-world settings with high-frame-rate video, large images, and a large number of noisy optical flow estimates. Our method runs in real-time on a single CPU and can estimate camera motion and scene depth in scenes with noisy optical flow with outliers, making it suitable for integration with filters for real-time navigation and for deployment on light-weight UAVs. The technical contributions of this paper are:
\begin{itemize}
	\item	
	A novel robust estimator based on the expected residual likelihood (ERL) of flow data that effectively attenuates the influence of outlier flow measurements and runs at 30-40 Hz on a single CPU.
	\item
	A novel robust optimization strategy using a lifted kernel that modifies the shape of the objective function to enable joint estimation of weights and model parameters, while enabling good empirical convergence properties.
\end{itemize}

\section{Related work}
\subsection{Egomotion/visual odometry}
Many approaches to the problem of visual odometry have been proposed. A distinction is commonly made between feature-based methods, which use a sparse set of matching feature points to compute camera motion, and direct methods, which estimate camera motion directly from intensity gradients in the image sequence. Feature-based approaches can again be roughly divided into two types of methods: those estimating camera motion from point correspondences between two frames (\textit{discrete} approaches) and those estimating camera motion and scene structure from the optical flow measurements induced by the motion between the two frames (\textit{continuous} approaches). In practice, point correspondences and optical flow measurements are often obtained using similar descriptor matching strategies. Nonetheless, the discrete and continuous approaches use different problem formulations, which reflect differing assumptions about the size of the baseline between the two camera positions. 

The continuous approach is the appropriate choice in situations where the real-world camera motion is slow relative to the sampling frequency of the camera. Our approach is primarily intended for situations in which this is the case, e.g. UAVs equipped with high-frame-rate cameras. Accordingly, we focus our review on continuous, monocular methods. For a more comprehensive discussion, see \cite{ma_invitation_2004}.

\subsection{Continuous, monocular approaches}

In the absence of noise, image velocities at 5 or 8 points can be used to give a finite number of candidate solutions for camera motion \cite{longuet-higgins_interpretation_1980} \cite{hartley_multiple_2003} \cite{nister_efficient_2004}. With more velocities, there is a unique optimal solution under typical scene conditions \cite{horn_motion_1988}. Many methods have been proposed to recover this solution, either by motion parallax \cite{longuet-higgins_interpretation_1980} \cite{hildreth_recovering_1992} \cite{heeger_subspace_1992} \cite{jepson_subspace_1990} or by using the so-called continuous epipolar constraint \cite{ma_invitation_2004}. The problem is nonlinear and nonconvex, but various linear approximation methods have been proposed to simplify and speed up estimation \cite{jepson_fast_1991} \cite{zhuang_simplified_1988} \cite{kanatani_3-d_1993}.

Although the problem has a unique optimum, it is characterized by many local minima, which pose difficulties for linear methods \cite{chiuso_optimal_2000}. Furthermore, in the presence of noise, many methods are biased and inconsistent in the sense that they do not produce correct estimates in the limit of an unlimited number of image velocity measurements \cite{zhang_consistency_2002}. Many methods also fail under many common viewing conditions or with a limited field of view \cite{daniilidis_analytical_1990}. Recently, \cite{fredriksson_fast_2014} and \cite{fredriksson_practical_2015} proposed branch-and-bound methods that estimate translational velocity in real time and effectively handle a large numbers of outliers. However, these methods deal with the case of pure translational camera motion, while our approach estimates both translational and rotational motion.

Most directly related to our work is the robust estimation framework presented in \cite{zhang_fast_1999}. They propose a method based on a variant of a common algebraic manipulation and show that this manipulation leads to an unbiased, consistent estimator. They pose monocular egomotion as a nonlinear least-squares problem in terms of the translational velocity. In this framework, angular velocity and inverse scene depths are also easily recovered after translational velocity is estimated. To add robustness, they use a loss function with sub-quadratic growth, which they solve by iteratively reweighted least squares (IRLS). We use a similar formulation but demonstrate several novel methods for estimating the parameters of a robust loss formulation. Our methods have properties that are well-suited for dealing with image sequences containing several thousand flow vectors in real time. In particular, we demonstrate that the ERL method adds robustness without requiring costly iterative reweighting, resulting in very little runtime overhead.

\begin{algorithm} [thpb]
	\scriptsize
	\centering
	\caption{ERL confidence weight estimation}
	\label{algorithm1}
	\begin{algorithmic}
		\renewcommand{\algorithmicrequire}{\textbf{Input:}}
		\renewcommand{\algorithmicensure}{\textbf{Output:}}
		\Require Measured flow $\{\vect{u}_n\}_{n=1}^{N}$, sampled translational velocities $\{\vect{t}_m\}_{m=1}^{M}$
		\Ensure Estimated confidence weights ${\{\hat{w}_n\}_{n=1}^N}$
		\ForAll {$m$}
		\State Compute scaled residuals:
		$$
		\vect{\tilde{r}}_{\vect{u}} = | A^{\perp}(\vect{t}_m)^{\top} (B\vect{\hat{\omega}}_m(\vect{t}_m) - \vect{u}) |		$$
		\State Compute maximum likelihood estimators of residual distribution:
		$$
		\hat{\mu}_m = \mathrm{median}(\tilde{r}_{\vect{u}}) 
		$$
		$$		
		\hat{b}_m = \frac{1}{N} \sum_{n=1}^{N}\| \tilde{r}_{\vect{u}_n} - \hat{\mu}_m \|
		$$
		\EndFor
		\ForAll {$n$}
		\State Compute confidence weights as expected likelihood under Laplacian fits:
		$$
		\hat{w}_n = \frac{1}{M} \sum_{m=1}^{M}\Lapl(\vect{\tilde{r}}_{\vect{u}_n}; \hat{\mu}_m, \hat{b}_m)
		$$
		\EndFor
		
		\Return ${\{\hat{w}_n\}_{n=1}^N}$
	\end{algorithmic}
\end{algorithm}

Other methods for monocular odometry augment velocity data with planar homography estimates \cite{geiger_stereoscan:_2011} \cite{song_parallel_2013} or depth filters \cite{forster_svo:_2014} to estimate scale. In this work, we do not rely on ground-plane estimation in order to maintain applicability to cases such as UAV navigation, where image sequences do not always contain the ground plane. Because we focus on frame-by-frame motion estimation, we cannot rely on a filtering approach to estimate depth. Our method can be augmented with domain-appropriate scale or depth estimators as part of a larger SLAM system.

%
%

\subsection{Robust optimization}
In this work, we propose to increase the robustness of monocular egomotion estimation (1) by estimating each flow vector's confidence weight as its expected residual likelihood (ERL) and (2) by using a lifted robust kernel to jointly estimate confidence weights and model parameters. ERL confidence weights are conceptually similar to the weights recovered in the IRLS method for optimizing robust kernels \cite{holland_robust_1977}. Robust kernel methods attempt to minimize the residuals of observations generated by the target model process ("inliers") while limiting the influence of other observations ("outliers"). Such methods have been used very successfully in many domains of computer vision \cite{geman_constrained_1992} \cite{black_unification_1996}. However, we are unaware of any previous work that attempts to estimate confidence weights based on the distribution of residuals at counterfactual model parameters, as we do in the ERL method.

The lifted kernel approach offers another method to design and optimize robust kernels in particularly desirable ways. Lifted kernels have recently been used in methods for bundle adjustment in SfM \cite{zach_robust_2014}, object pose recovery \cite{zach_dynamic_2015}, and non-rigid object reconstruction \cite{zollhofer_real-time_2014}. Our lifted kernel approximates the truncated quadratic loss, which has a long history of use in robust optimization in computer vision \cite{blake_visual_1987} and has demonstrated applicability in a wide variety of problem domains.

Previous studies have used robust loss functions for monocular egomotion \cite{zhang_fast_1999}, visual SLAM \cite{newcombe_dtam:_2011}, and RGB-D odometry \cite{kerl_robust_2013}. To our knowledge, we present the first application of lifted kernels for robust monocular egomotion. Noise is typically handled in odometry by using sampling-based iterative methods such as RANSAC, which makes use of a small number of points to estimate inlier sets (typically five or eight points in monocular methods). The use of a robust kernel allows us to derive our final estimate from a larger number of points. This is desirable because the structure of the problem of continuous monocular odometry admits fewer correct solutions when constrained by a larger number of input points, which can better reflect the complex depth structure of real scenes. Our robust methods allow us to take advantage of a large number of flow estimates, which, while noisy, may each contribute weakly to the final estimate.

\section{Problem formulation and approach}

In this section, we present the continuous formulation of the problem of monocular visual egomotion. We describe and motivate our approach for solving the problem in the presence of noisy optical flow. We then describe two methods for estimating the confidence weights for each flow vector in a robust formulation of the problem, as well as the pipeline we use to estimate camera motion and scene depth.

\subsection{Visual egomotion computation and the motion field}

In the continuous formulation, visual egomotion methods attempt to estimate camera motion and scene parameters from observed local image velocities (optical flow). The velocity of an image point due to camera motion in a rigid scene under perspective projection is given by 
$$
\vect{u}(\vect{x}_i) = \rho(\vect{x}_i)A(\vect{x}_i)\vect{t} + B(\vect{x}_i)\vect{\omega}. \eqno{(1)}
$$
where $\vect{u}_i(\vect{x}_i)=(u_i,v_i)^\top \in \bR^2$ is the velocity (optical flow) at image position $\vect{x}_i=(x_i,y_i)^\top\in \bR^2$, $\vect{t}=(t_x,t_y,t_z)^\top \in \bR^3$ is the camera's instantaneous translational velocity, $\vect{\omega}=(\omega_x, \omega_y, \omega_z)^\top \in \bR^3$ is the camera's instantaneous rotational velocity, and $\rho(\vect{x}_i)=\frac{1}{Z(\vect{x_i})} \in \bR$ is the inverse of scene depth at $\vect{x}_i$ along the optical axis. We normalize the camera's focal length to 1, without loss of generality. In the case of calibrated image coordinates,
$$
A(\vect{x}_i) = \begin{bmatrix} 1 && 0 && -x_i \\ 0 && 1 && -y_i \end{bmatrix},
$$
$$
B(\vect{x}_i) = \begin{bmatrix} -x_iy_i && 1+x_i^2 && -y_i \\ -1-y_i^2 && x_iy_i && x_i \end{bmatrix}.
$$

This formulation is appropriate for the small-baseline case where point correspondences between frames can be treated as 2D motion vectors.

The goal of monocular visual egomotion computation is thus to estimate the six motion parameters of $\vect{t}$ and $\vect{\omega}$ and the $N$ values for $\rho$ from $N$ point velocities $\vect{u}$ induced by camera motion. $\vect{t}$ and $\rho$ are multiplicatively coupled in equation (1) above, so $\vect{t}$ can only be recovered up to a scale. We therefore restrict estimates of $\vect{t}$ to the unit hemisphere, $\|\vect{t}\|=1$.

The full expression for the set of $N$ point velocities can be expressed compactly as
$$
\vect{u} = A(\vect{t})\vect{\rho} + B\vect{\omega}. \eqno{(2)}
$$
where the expressions for $A(\vect{x})$, $B(\vect{x})$, and $\rho(\vect{x})$ for all $N$ points are
$$
A(\vect{t}) = \begin{bmatrix}
    	A(\vect{x}_1)\vect{t} &    0    & \ldots &    0    \\
    	0   & A(\vect{x}_2)\vect{t} & \ldots &    0    \\
    	\vdots &  \vdots & \ddots & \vdots  \\
    	0   &     0   & \ldots & A(\vect{x}_N)\vect{t}
    \end{bmatrix} \in \bR^{2N \times N}
$$
$$
    B =  \begin{bmatrix}
    	B(\vect{x}_1) \\
    	B(\vect{x}_2) \\
    	\vdots \\
    	B(\vect{x}_N) 
    \end{bmatrix} \in \bR^{2N \times 3}
$$

\noindent and the velocity and depth for each of the points are concatenated to form \(\vect{u} = (\vect{u}_1^{\top}, \vect{u}_2^{\top}, \ldots, \vect{u}_N^{\top})^{\top} \in \bR^{2N \times 1}\) and \(\vect{\rho} = (\rho(\vect{x}_1), \rho(\vect{x}_2), \ldots, \rho(\vect{x}_N))^{\top} \in \bR^{N \times 1} \). We estimate camera motion and scene depth by minimizing the objective
$$ 
\min_{\vect{t},\vect{\rho},\vect{\omega}} E(\vect{t},\vect{\rho},\vect{\omega}) = \min_{\vect{t},\vect{\rho},\vect{\omega}}L(r(\vect{t},\vect{\rho},\vect{\omega}))
$$
$$
= \min_{\vect{t},\vect{\rho},\vect{\omega}} \| A(\vect{t})\vect{\rho} + B\vect{\omega} - \vect{u} \|_2^2. \eqno{(3)}
$$
Here, $L(x):\bR^N \rightarrow \bR$ is a loss function and $r(\vect{t},\vect{\rho},\vect{\omega}):\bR^{N+6} \rightarrow \bR^N$ is a residual function for the flow field depending on the estimated model parameters. We first describe the case of an unweighted residual function under a quadratic loss, which is suitable for the case of Gaussian noise.

Following \cite{zhang_fast_1999}, we note that no loss of generality occurs by first solving this objective for $\rho$ in the least-squares sense. Minimizing over $\rho$ gives
$$
\min_{\vect{t},\vect{\omega}} \min_{\vect{\rho}} \| A(\vect{t})\vect{\rho} + B\vect{\omega} - \vect{u} \|_2^2 \\
$$
$$
= \min_{\vect{t},\vect{\omega}} \| A^{\perp}(\vect{t})^{\top} (B\vect{\omega} - \vect{u}) \|_2^2, \eqno{(4)}
$$
where $A^{\perp}(\vect{t})$ is the orthogonal compliment to $A(\vect{t})$. This expression no longer depends on $\vect{\rho}$ and depends on $\vect{t}$ only through $A^{\perp}(\vect{t})^\top$, which is fast to compute due to the sparsity of $A(\vect{t})$ (see section II of the supplement for more details).

In the absence of noise, we could proceed by directly minimizing equation (4) in $\vect{t}$ and $\vect{\omega}$. In particular, given a solution for $\vect{t}$, we can directly solve for $\vect{\omega}$ by least squares in $O(N)$ time. In the noiseless case, we estimate $\vect{t}$ by optimizing
$$
\min_{\vect{t}} \| A^{\perp}(\vect{t})^{\top} (B\hat{\vect{\omega}}(\vect{t}) - \vect{u}) \|_2^2, \eqno{(5)}
$$
where $\hat{\vect{\omega}}(\vect{t})$ is the least-squares estimate of $\vect{\omega}$ for a given $\vect{t}$ (see section IV of the supplement for more details). This method of estimating $\vect{t}$, $\vect{\rho}$, and $\vect{\omega}$ was shown to be consistent in \cite{zhang_consistency_2002}. That is, in the absence of outliers, this method leads to arbitrarily precise, unbiased estimates of the motion parameters as the sample size increases.

\subsection{Robust formulation}
However, the manipulations introduced in equations (4) and (5) rely on least-squares solutions and are not stable in the presence of outliers.  Accordingly, instead of directly solving (5), we propose to solve a robust form. To do so, we introduce a confidence weight for each flow vector $w_i(\vect{u_i}) \in [0,1]$ to give
$$
\min_{\vect{t}} L(r(\vect{t},\vect{\hat{\omega}}(\vect{t})), \vect{w})
$$
$$
= \min_{\vect{t}} \| \vect{w} \circ A^{\perp}(\vect{t})^{\top} (B\hat{\vect{\omega}}(\vect{t}) - \vect{u}) \|_2^2, \eqno{(6)}
$$
where $\vect{w} =(w(\vect{u}_1), w(\vect{u}_2), \ldots, w(\vect{u}_N))^{\top} \in [0,1]^N $ is the vector of all weights, $\vect{r} \in \bR^N$ is the vector of residuals for the flow field at some estimate of $\vect{t}$, and $\circ$ is the Hadamard product.

Each entry $w(\vect{u}_i)$ of $\vect{w}$ attempts to weight the corresponding data point $\vect{u}_i$ proportionally to its residual at the optimal model parameters ($\hat{\vect{t}}, \hat{\vect{\rho}}, \hat{\vect{\omega}}$), reflecting the degree to which the point is consistent with a single generating function for the motion in the scene, possibly with Gaussian noise. In other words, it reflects the degree to which $\vect{u}_i$ is an inlier for the optimal model of camera motion in a rigid scene. This is equivalent to replacing the choice of $L(x)=x^2$ as the loss in equation (5) with a function that grows more slowly.

We introduce a method to directly estimate the confidence weights as the expected residual likelihood (ERL) for each flow vector given the distribution of residuals for the flow field at a range of model parameters consistent with the solution in (5). We interpret each weight in terms of an estimate of the validity of the corresponding point under the model: that is, as an estimate of the point's residual at the optimal model parameters in a noise-free context. We compare ERL to a method that replaces $L(x)=x^2$ in (5) with a lifted truncated quadratic kernel \cite{zach_robust_2014} and jointly optimizes the confidence weights and model parameters. We demonstrate that ERL outperforms the lifted kernel approach on the KITTI dataset, and both of these approaches outperform existing methods for monocular egomotion computation.

\subsection{Confidence weight estimation by expected residual likelihood}

Here, we describe the ERL method for estimating the confidence weights in (6), and we demonstrate that this method provides a good estimate of the appropriate confidence weights in the case of optical flow for visual egomotion.

At the optimal model parameters, $(\vect{t}^*, \vect{\rho}^*, \vect{\omega}^*)$, the residuals for inlier points (i.e. correct flow vectors due to rigid motion) are distributed according to a normal distribution, reflecting zero-mean Gaussian noise. However, in the presence of outliers, a zero-mean Laplacian distribution provides a better description of the residual distribution (see \textbf{Supplemental Fig. 2}). Accordingly, we can fit a Laplacian distribution to the observed residuals at the optimal model parameters to approximate the probability density function for residuals.

We use this property to identify outliers as those points that are inconsistent with the expected residual distribution at a range of model values. For each point, we compute the likelihood of each observed, scaled residual as
$$
p(\tilde{r}^m_{\vect{u}_i} | (\vect{t}_m, \vect{\rho}_m, \vect{\omega}_m), \tilde{\vect{r}}^m_{\vect{u}}) = \Lapl(\tilde{r}_{\vect{u}_i}; \hat{\mu}_m,\hat{b}_m), \eqno{(7)}
$$
where $\tilde{r}^m_{\vect{u}_i}$ is the scaled residual under the m\textsuperscript{th} model $(\vect{t}_m, \vect{\rho}_m, \vect{\omega}_m)$ at the i\textsuperscript{th} flow vector and $\tilde{\vect{r}}^m_{\vect{u}}=(\tilde{r}^m_{\vect{u}_1},\tilde{r}^m_{\vect{u}_2},...,\tilde{r}^m_{\vect{u}_N})^\top$. We fit $\hat{\mu}_m$ and $\hat{b}_m$, respectively the location and scale parameters of the Laplacian distribution, to the set of scaled residuals $\tilde{\vect{r}}^m_{\vect{u}}$ using maximum likelihood.

\begin{figure}[thpb]
	\centering
	\includegraphics[scale=0.8]{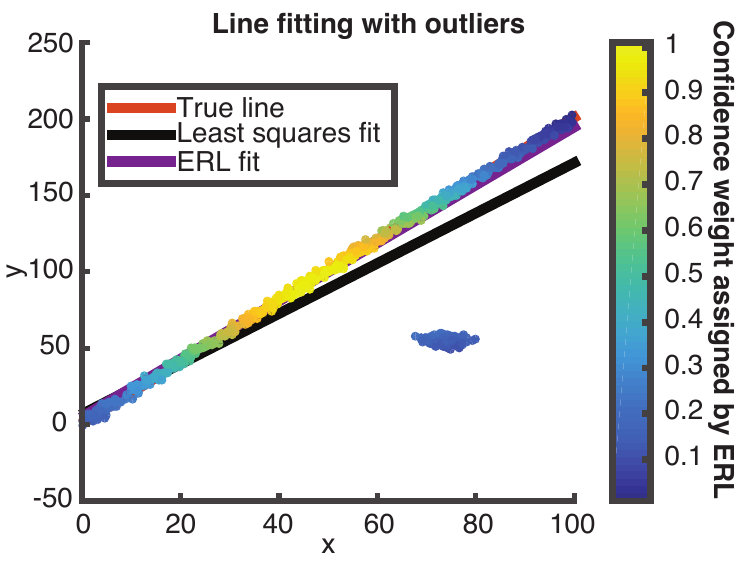}
	%
	\caption{A 2D line-fitting problem demonstrating how ERL weights inliers and outliers. Inliers are generated as $y_i \approx 2x_i+1$ with Gaussian noise. Each data points is colored according to its estimated confidence weight.}
	\label{figurelabel}
\end{figure}

Because inliers exhibit smaller self-influence than outliers \cite{huber_robust_2011}, inlier residuals will typically be associated with higher likelihood values. However, the distribution used to estimate the likelihood reflects both the inlier and outlier points. If the counterfactual model parameters used to estimate the m\textsuperscript{th} likelihood correspond to a model that is highly suboptimal, some outliers may be assigned higher likelihoods than they would be at the optimal model. Moreover, the presence of Gaussian noise means that the estimated likelihood for individual inliers may be erroneously low by chance for a particular model even if the optimal exponential distribution is exactly recovered. 

To arrive at more reliable estimates and to discount the effect of erroneous likelihoods due to the specific model parameters being evaluated, we estimate the expected residual likelihood for each data point by evaluating the likelihood under $M$ models,
$$
\hat{w}_i = \bE[\tilde{r}^m_{\vect{u}_i}] = \frac{1}{M} \sum_{m=1}^{M}\Lapl(\vect{\tilde{r}}_{\vect{u}_i}; \hat{\mu}_m, \hat{b}_m). \eqno{(8)}
$$
This method returns a vector $\hat{\vect{w}} \in \bR^N$. To use $\hat{\vect{w}}$ as confidence weights in a robust optimization context, we scale them to the interval $[0,1]$. Scaling the maximum $\hat{w}_i$ to 1 and the minimum $\hat{w}_i$ to 0 for each flow field works well in practice.

\begin{figure}[thpb]
	\centering
	\includegraphics[scale=0.35]{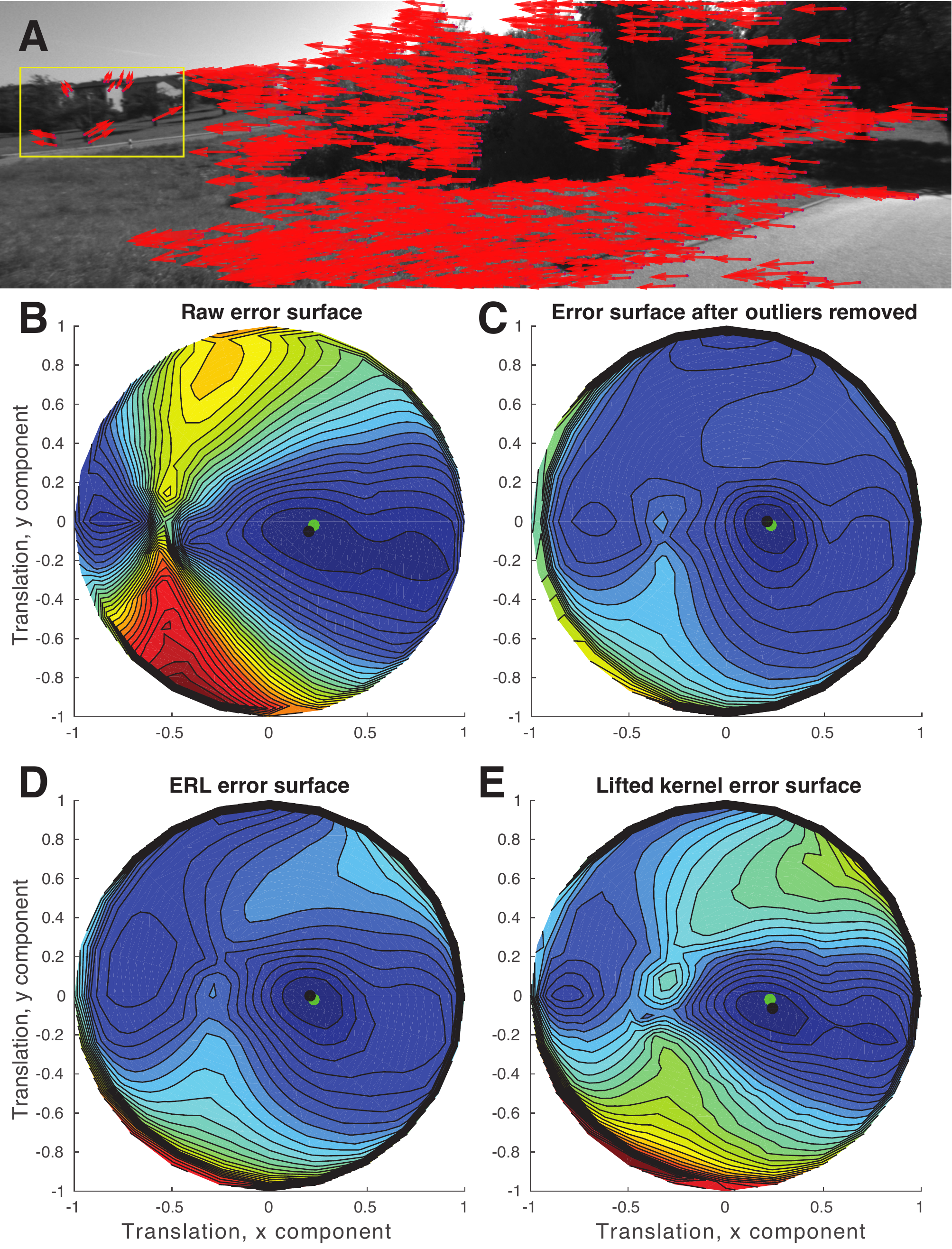}
		\caption{Robust methods recover the error surface of the outlier-free flow field. (A) Example optical flow field from two frames of KITTI odometry (sequence 10, images 14-15). Note the prominent outliers indicated by the yellow box. Error surfaces on this flow field for (A) the raw method (equation (5)) with all flow vectors, (B) with outliers removed by hand, and (C) with confidence weights estimated by ERL or (D) the lifted kernel. The green point is the true translational velocity and the black point the method's estimate. Blue: low error. Red: high error. Translation components are given in calibrated coordinates.}
		\label{figurelabel}
	\end{figure}

The full process to estimate weights by ERL is shown in \textbf{Algorithm 1}. This method returns confidence weights in $O(MN)$ time, where $M$ is set by the user. Empirically, the ERL method gives results that reflect the inlier structure of the data with small values of $M$ (we use $M \approx 100$), allowing very quick runtimes. In practice, the method assigns high weights to very few outliers while assigning low weights to acceptably few inliers. Thus, the method balances a low false positive rate against a moderately low false negative rate. This is a good strategy because our method takes a large number of flow vectors as input, which leads to redundancy in the local velocity information. \textbf{Fig. 2} illustrates the ERL method's use in a simple 2D robust line-fitting application.

As discussed above, choosing values for the confidence weights in a least squares objective is equivalent to fitting a robust kernel. We note that regression under the assumption of Laplacian noise leads to an L1 cost. However, we have no guarantees about the form of the robust kernel corresponding to the weights chosen by the ERL method. Accordingly, we also explored using a robust kernel with known properties.

\subsection{Robust estimation using a lifted kernel}

Here, we explore the effect of jointly optimizing the confidence weights, $\vect{w}(\vect{u})$, and $\vect{\omega}$ for a given value of $\vect{t}$ using the lifted kernel approach described in \cite{zach_robust_2014}. In our case, a lifted kernel takes the form 
$$ 
\min_{\vect{t},\vect{\omega},\vect{w}}\hat{L}(r(\vect{t},\vect{\omega}),\vect{w})
$$
$$
=
\min_{\vect{t}}\min_{\vect{\omega},\vect{w}} (\| \vect{w} \circ A^{\perp}(\vect{t})^{\top} (B\vect{\omega}(\vect{t}) - \vect{u}) \|^2_2 + \sum_{i=1}^N \kappa^2(w_i^2)), \eqno{(9)}
$$
where the lifted kernel of the loss $L$ is denoted as $\hat{L}$. $\kappa(x):\bR \rightarrow \bR$ is a regularization function applied to the weights. Because this approach does not rely on the least squares solution for rotational velocity, $\vect{\hat{\omega}}$, it may gain additional robustness to noise. This approach also allows us to estimate the confidence weights for particular values of $\vect{t}$, unlike the ERL approach, which relies on estimates at several values of $\vect{t}$ to produce stable results.

Different choices of $\kappa$ produces different kernels. We use
$$
\kappa(w^2) = \frac{\tau}{\sqrt{2}}(w^2-1), \eqno{(10)}
$$ 
which gives a kernel that is a smooth approximation to the truncated quadratic loss \cite{zach_robust_2014}. $\tau$ is a hyperparameter that determines the extent of the quadratic region of the truncated quadratic loss. We set $\tau=0.05$ for all results shown here, but other choices give similar results.

\begin{figure}[thpb]
	\centering
	\includegraphics[scale=0.33]{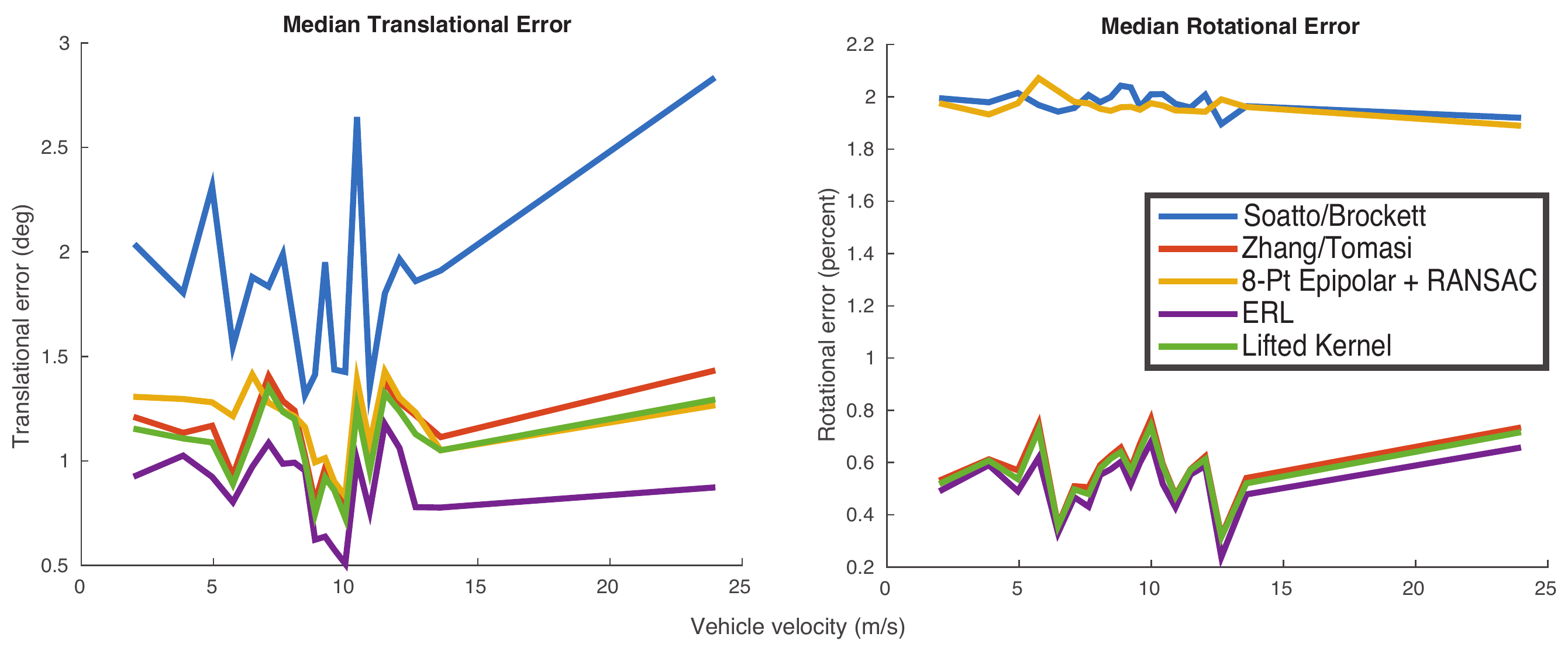}
	\caption{Median translational and rotational errors on the full KITTI odometry dataset for our methods and baselines.}
	\label{figurelabel}
\end{figure}			

The lifted kernel approach to solving nonlinear least squares problems is similar to IRLS insofar as it incorporates confidence weights on each of the data points and optimizes the values of these weights in addition to the value of the target model parameters. However, rather than alternately estimating the best weights given estimated model parameters and the best model parameters given estimated weights, the lifted approach simultaneously optimizes for both weights and model parameters, effectively "lifting" a minimization problem to a higher dimension.

The lifted kernel approach has several properties that are particularly beneficial for encouraging fast convergence. First, by using the weights to increase the dimensionality of the optimization problem, the lifted kernel minimizes the extent of regions of low gradient in the cost function. This ensures the method can quickly and reliable converge to minima of the function. Second, optimization can exploit the Gauss-Newton structure of the joint nonlinear least-squares formulation for faster convergence than the slower iterative-closest-points-like convergence exhibited by IRLS.

To illustrate the effect of our two robust optimization strategies, we display the error surfaces for the ERL and lifted-kernel methods on a sample flow field from KITTI (\textbf{Fig. 3}). The error surfaces are shown as a function of the translational velocity. Both methods recover error surfaces that resemble the error due to inlier flow vectors. The confidence weights estimated by ERL generally more closely resemble the pattern of inliers and outliers in flow data. To produce the results for the case with outliers removed, we strengthened the maximum bidirectional error criterion for flow inclusion to eliminate noisy matches and manually removed obvious outliers from the flow field.

\section{Experiments}
We compare the performance of the proposed methods (called "ERL" and "Lifted Kernel" in the figures) to several baseline methods for monocular egomotion/visual odometry from the literature: 5-point epipolar+RANSAC (using \cite{stewenius_recent_2006}), 8-point epipolar+RANSAC (using \cite{Corke11a}), and two continuous epipolar methods - Zhang/Tomasi \cite{zhang_fast_1999}, which is identical to equation (5), and Soatto/Brockett \cite{soatto_optimal_1998}. All experiments were run on a desktop with an Intel Core i7 processor and 16 GB of RAM. A single CPU core was used for all experiments. 

With $\sim$1000 flow vectors, the ERL method runs at 30-40 Hz in an unoptimized C++ implementation. Because of the low overhead of the ERL procedure, this is effectively the same runtime as the Zhang/Tomasi method. The lifted kernel optimization has no convergence guarantees, and it typically runs at \textless 1 Hz in a MATLAB implementation. Note that both of these runtimes can be significantly improved with better optimization. The Soatto/Brockett method runs extremely quickly (\textgreater 500 Hz), but performs poorly on real sequences. The implementation of epipolar+RANSAC used here runs at $\sim$25 Hz. Optical flow for all our results was extracted using a multiscale implementation of the KLT method \cite{lucas_iterative_1981} \cite{tomasi_detection_1991}.

For both ERL and the lifted approach, we optimize $\vect{t}$ using Gauss-Newton. We initialize $\vect{t}$ at a grid of values spaced over the unit hemisphere to decrease the chance of converging to a non-global minimum. We then prune the grid to a single initial value $\vect{t}_0$ by choosing the grid point that gives the lowest residual under equation (6) or (9) for ERL or the lifted kernel, respectively. We then optimize to convergence starting from $\vect{t}_0$. This pruning strategy is effective at avoiding local minima because good estimates for the weights return an error surface that is very similar to the noiseless case (see \textbf{Fig. 3}) and this error surface is smooth with respect to the sampling density we use (625 points) \cite{chiuso_optimal_2000}. Confidence weights for ERL are computed using model parameters sampled on a coarser grid (100 points), as this is adequate to give good confidence weight estimates.

For all tests using the lifted kernel, we optimize the expression in equation (9) using the efficient Schur compliment implementation of Levenberg-Marquardt described in \cite{zach_robust_2014}. Details of the optimization procedure used here are given in section III of the supplement. We did not explore jointly optimizing over $\vect{t}$, $\vect{\omega}$, and $\vect{w}$, but joint optimization over these model parameters with a lifted kernel is possible, and we plan to explore its use in future work. 

\subsection{Evaluation on KITTI}
We evaluate the performance of our method using the KITTI dataset \cite{geiger_are_2012}, which is a collection of real-world driving sequences with ground-truth camera motion and depth data. The sequences contained in the dataset are challenging for state-of-the-art odometry methods for several reasons. First, they contain large inter-frame motions and repetitive scene structures that make estimating accurate flow correspondences difficult in real time. Second, several sequences feature little to no camera motion, which typically causes monocular odometry methods to fail. Finally, some sequences contain independent motion due to other vehicles and pedestrians, which violates the assumption of scene rigidity and makes reliable odometry more difficult.

\begin{figure}[th!pb]
	\centering
	\includegraphics[scale=0.45]{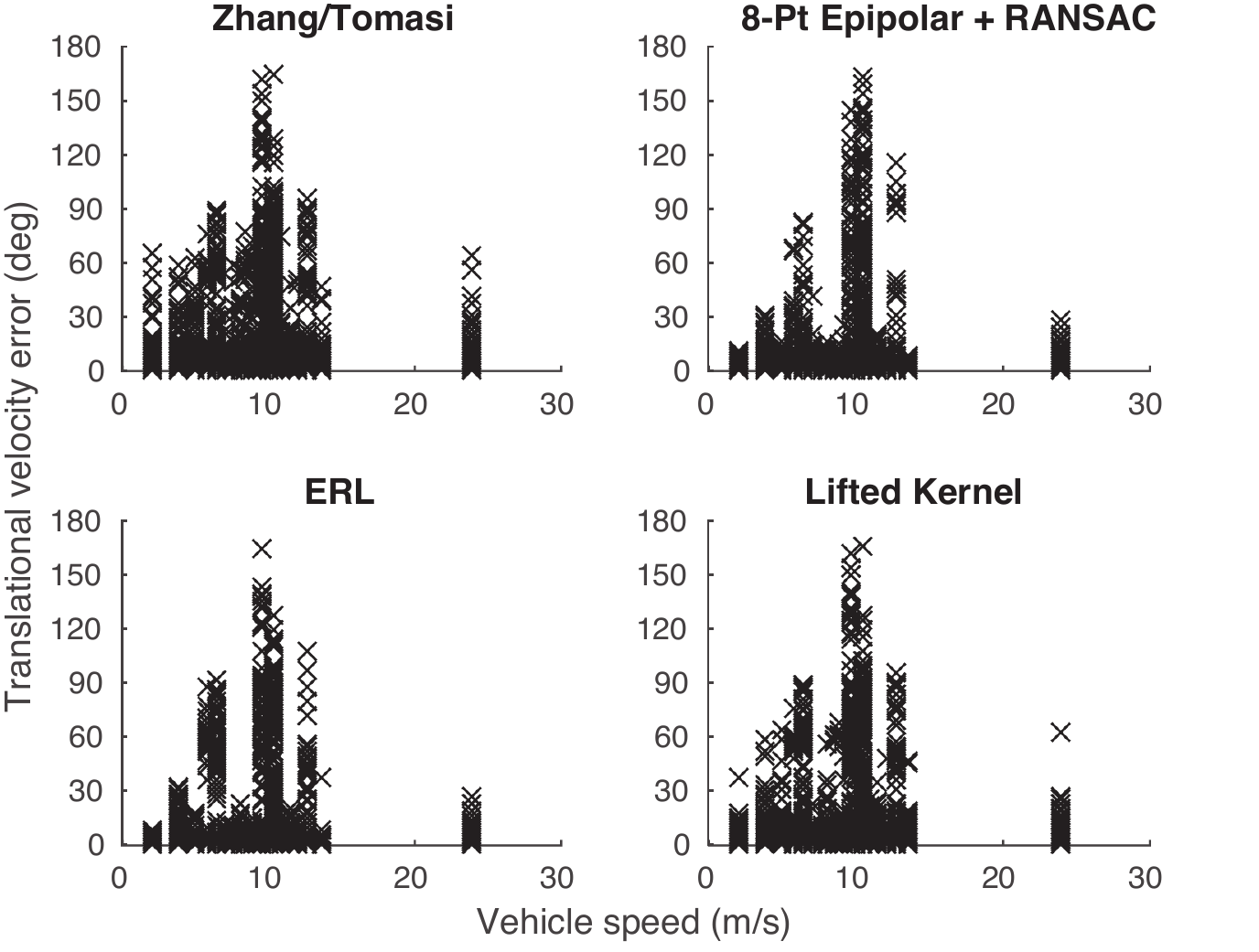}	
	\caption{Full distribution of translational velocity errors.}
	\label{figurelabel}
\end{figure}

\begin{figure}[th!pb]
	\centering
	\includegraphics[scale=0.45]{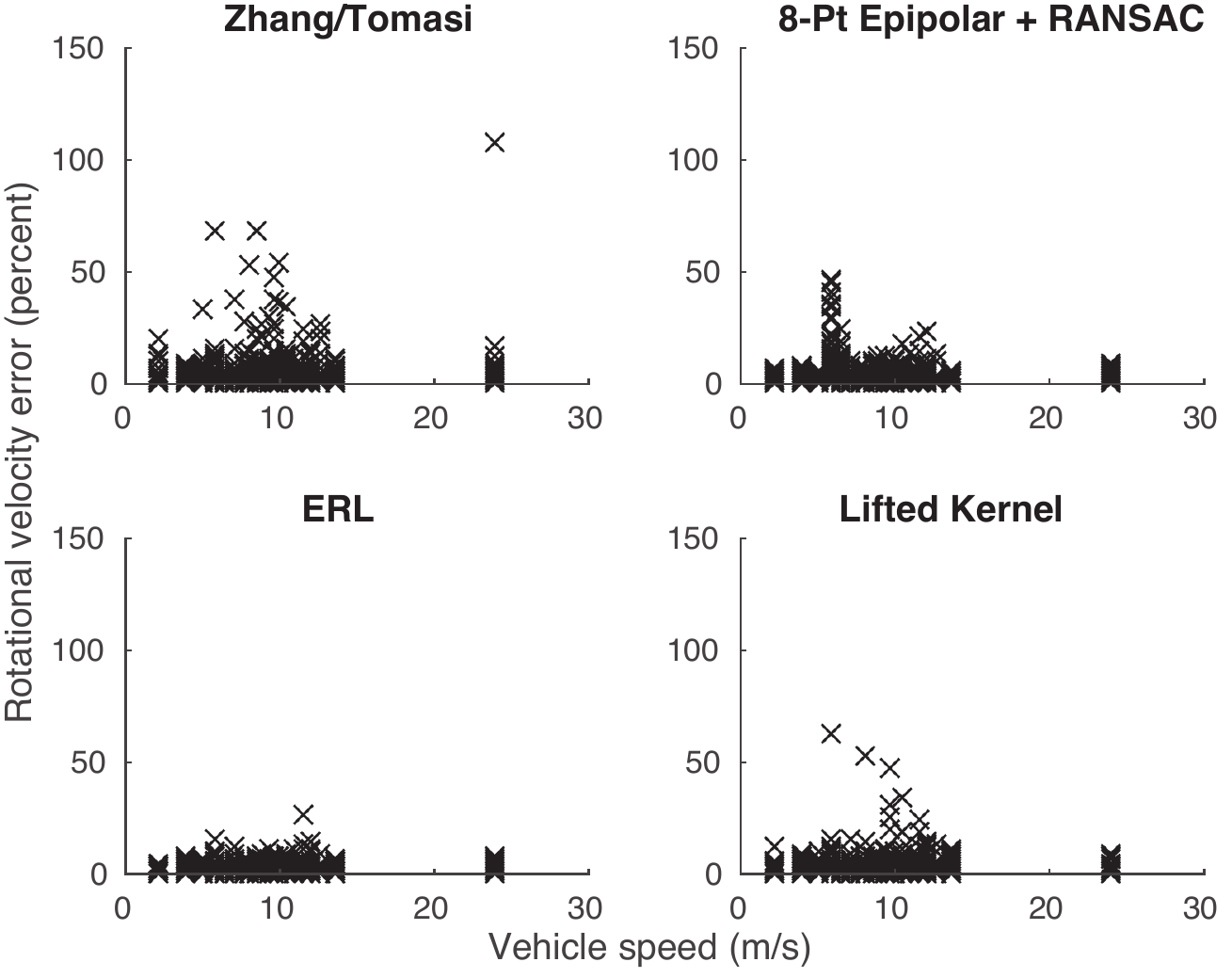}	
	\caption{Full distribution of rotational velocity errors.}
	\label{figurelabel}
\end{figure}

All results are performed on neighboring frames of the KITTI odometry dataset (no skipped-frame sequences are evaluated), as these image pairs better match the modeling assumptions of continuous egomotion/odometry methods. All sequences were captured at 10 Hz at a resolution of 1392 x 512 pixels. We evaluated all methods on all 16 sequences of the KITTI odometry test set.

The results for methods on KITTI are shown in \textbf{Figs. 4-6}. For ease of visualization, the results for the 5-point epipolar method with RANSAC are not shown (they were significantly worse than all other methods we attempted). ERL produces the best estimates of translational velocity, while the lifted kernel produces results of similar quality to 8-point epipolar with RANSAC and the Zhang/Tomasi method. ERL, the lifted kernel, and Zhang/Tomasi produce rotational velocity estimates of similar quality. The 8-point epipolar method produces worse estimates in this case because of the large baseline assumption, which is not suitable for rotational velocity estimation under these conditions. Soatto/Brockett produces bad estimates in these test cases because of the bias introduced by its algebraic manipulation.

\subsection{Synthetic sequences}
To estimate the robustness of our methods to outliers, we test the methods on synthetic data. Synthetic data were created by simulating a field of 1500 image points distributed uniformly at random depths between 2 and 10 m in front of the camera and uniformly in $x$ and $y$ throughout the frame. A simulated camera is moved through this field with a translational velocity drawn from a zero-mean Gaussian with standard deviation of 1 m/frame and a rotational velocity drawn from a zero-mean Gaussian with standard deviation of 0.2 radians/frame. Flow was generated from the resulting 3D point trajectories by perspective projection using a camera model with a 1 m focal length. All flow vectors were corrupted with noise in a random direction and magnitude drawn from a zero-mean Gaussian with a standard deviation 1/10\textsuperscript{th} the mean flow vector magnitude. Outliers were created by replacing a fraction of the points with random values drawn from a Gaussian fit to the magnitude and direction of all inlier flow vectors. We ran 100 iterations at each outlier rate. We ran all egomotion methods on the same data.

\begin{figure}[th!pb]
	\centering
	\includegraphics[scale=0.4]{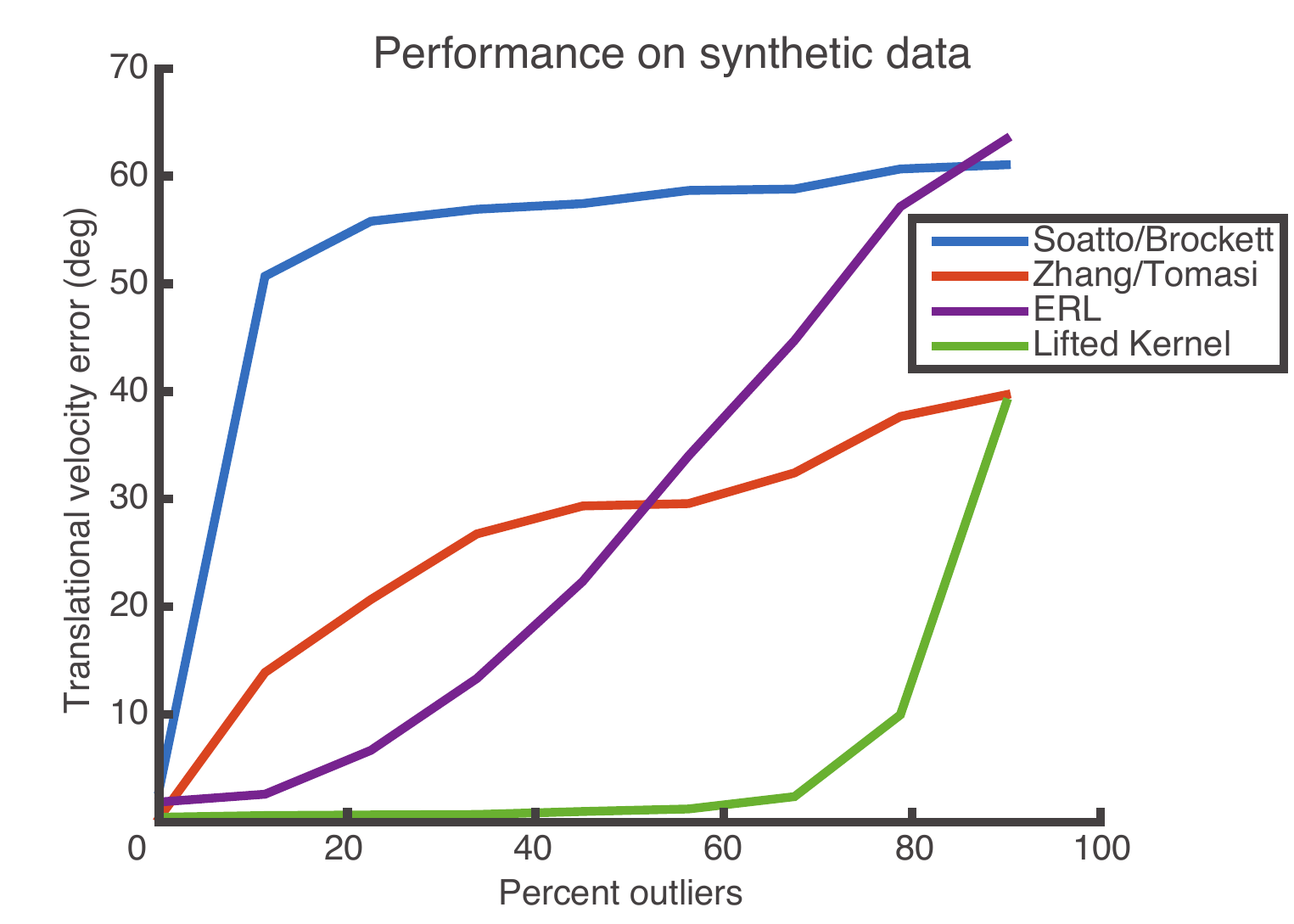}
	\caption{Translation error as a function of percent outliers on synthetic data for our robust methods and two baseline continuous egomotion methods.}
	\label{figurelabel}
\end{figure}	

The errors in translational motion estimated on this data are shown in \textbf{Fig. 7}. As expected, the two robust methods outperform least-squares methods for reasonable numbers of outliers. At higher outlier rates, however, the performance of both robust methods deteriorates. Interestingly, the performance of the lifted kernel method is stable even when the majority of data points are outliers. We are uncertain why the lifted kernel performs better than ERL on synthetic data, while the opposite is true for KITTI. This difference may be due to the way the data were generated - in KITTI, outliers often reflect real structures in the scene and may contain some information about camera motion, but this is not the case in the synthetic data. The difference may also be due in part to the difference in depth structures in KITTI and the synthetic data. In KITTI, flow magnitude for both inliers and outliers is reflective of depth structure, and depth in real scenes is not distributed uniformly.

\section{Conclusions}

We have introduced new techniques for robust, continuous egomotion computation from monocular image sequences. We described ERL, a novel robust method that directly estimates confidence weights for the vectors of a flow field by evaluating the distribution of flow residuals under a set of self-consistent counterfactual model parameters. We also introduced a new formulation of the perspective motion equation using a lifted kernel for joint optimization of model parameters and confidence weights. We compared the results of ERL and the lifted kernel formulation, and showed that while the lifted kernel appears to be more stable in the presence of a large fraction of outliers, ERL performs better in a real-world setting. The ERL method achieves good results on KITTI without relying on stereo data or ground plane estimation and accordingly is well-suited for use in lightweight UAV navigation. We are unable to directly evaluate our methods on this target domain because there are currently no UAV datasets with suitable ground truth. Although the empirical results here are promising, we have no guarantees on the weights recovered by ERL, and this remains a topic for future work. 

Our code is publicly available at \url{https://github.com/stephenphillips42/erl_egomotion}.

\section*{Acknowledgments}

The authors gratefully acknowledge support by the grants NSF-DGE-0966142, NSF-IIP-1439681, NSF-IIS-1426840, ARL MAST-CTA W911NF-08-2-0004, and ARL RCTA W911NF-10-2-0016.

{\small
\bibliographystyle{IEEEtran}
\bibliography{root}
}

\pagebreak
\section*{Supplemental material}
\beginsupplement

\section{Supplemental Experiments}
\subsection{Goodness of Fit for Laplacian Distribution}
To justify the use of a Laplacian distribution for ERL, we used ground truth flow fields to examine the distribution of errors in estimated optical flow. Ground truth flow was obtained using the KITTI Stereo dataset. Flow was produced according to equation (1) of the main paper. All images containing both depth and odometry ground truth were used. Errors were obtained for flow at all points that both contained ground truth depth and produced a sufficiently good KLT flow vector, using the same inclusion criteria as the main paper. We fit Laplacian and Gaussian distributions to the errors in the estimated optical flow, and computed the sum of the log likelihoods of each errors in the estimated distributions. In \textbf{Supplemental Figure \ref{fig:laplacianfit}} we plot the relative likelihoods of the data under the two distributions, and it is clear that the Laplacian fits consistently produce a higher likelihood than the Gaussian fits.
\begin{figure}[thpb]
	\centering
	\framebox{\parbox{3in}{
			\includegraphics[scale=0.5]{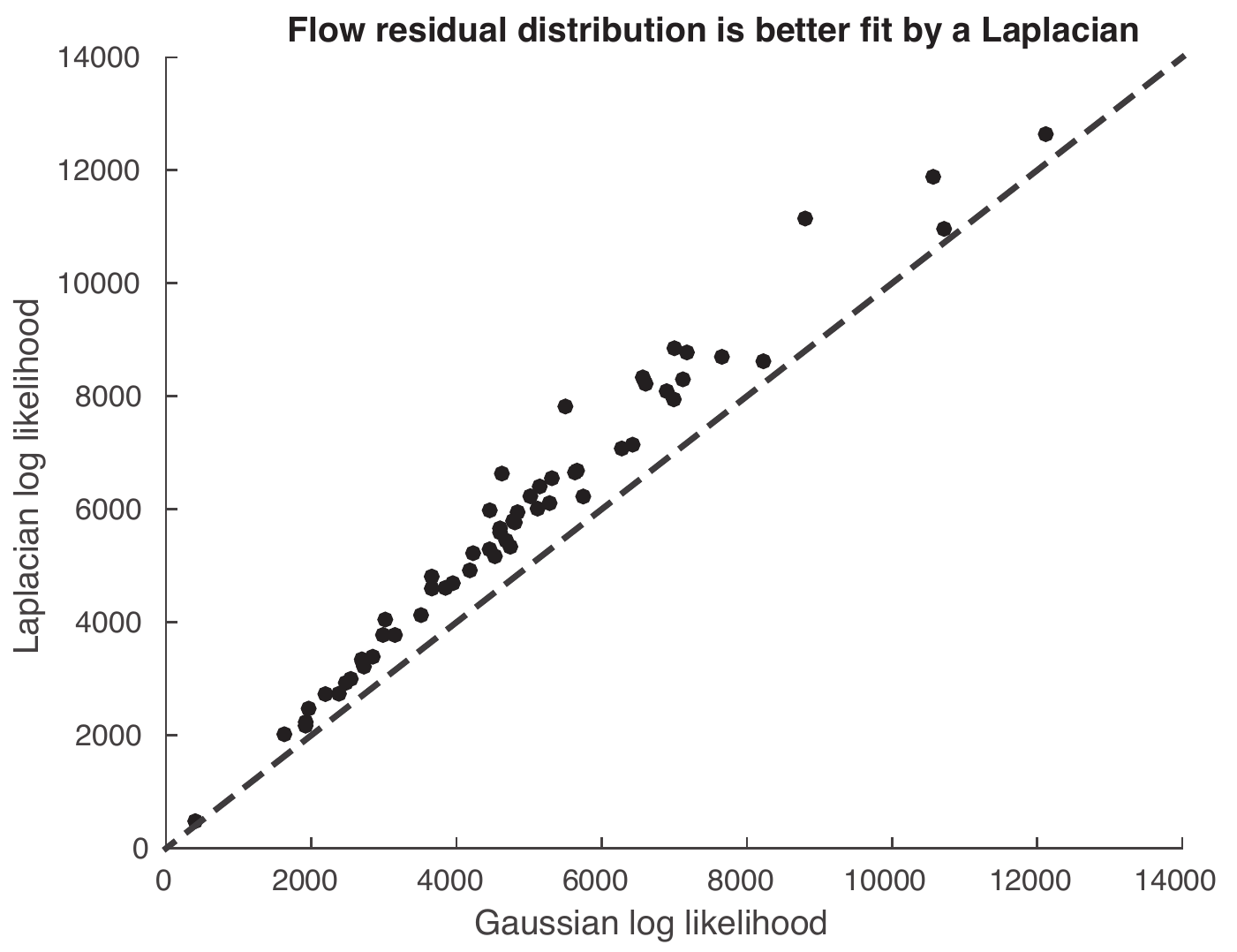}
		}}	
		\caption{Log likelihoods of Laplacian and Gaussian fits to the optical flow error. Laplacian fits are consistently better than Gaussian fits.}
		\label{fig:laplacianfit}
	\end{figure}
	
	\subsection{Comparison Of Difference ERL Likelihood Schemes}
	We also compared the results obtained by ERL different likelihood functions over the KITTI Odometry dataset. We compare the results obtained using a Laplacian or a Gaussian fit to compute the weights in ERL. Results are shown in \textbf{Supplemental Figure \ref{fig:likelihoodschemes}}. The use of a Laplacian distribution leads to a small but consistent improvement.
	
	\begin{figure}[thpb]
		\centering
		\framebox{\parbox{3in}{
				\includegraphics[scale=0.5]{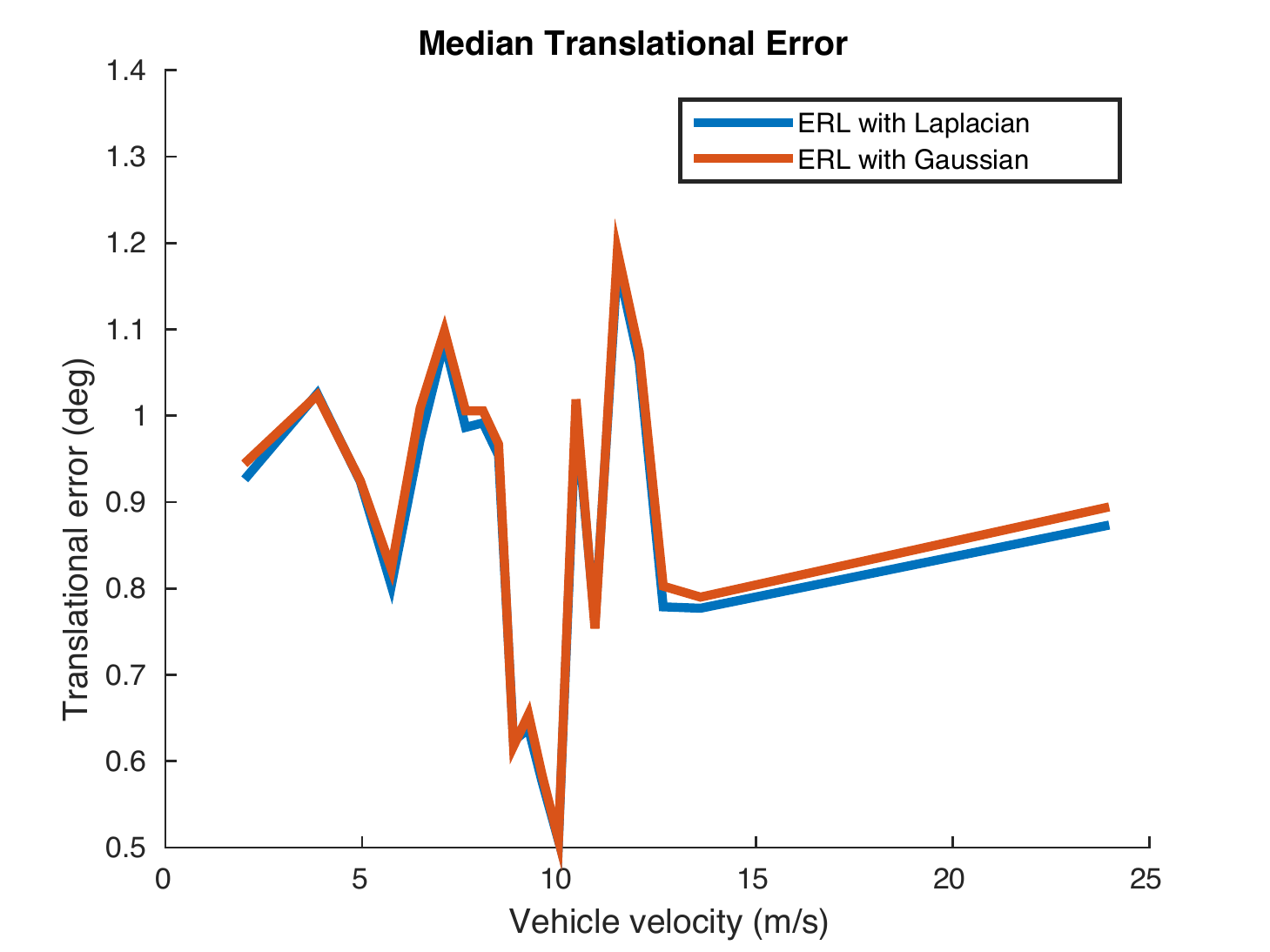}
				\includegraphics[scale=0.5]{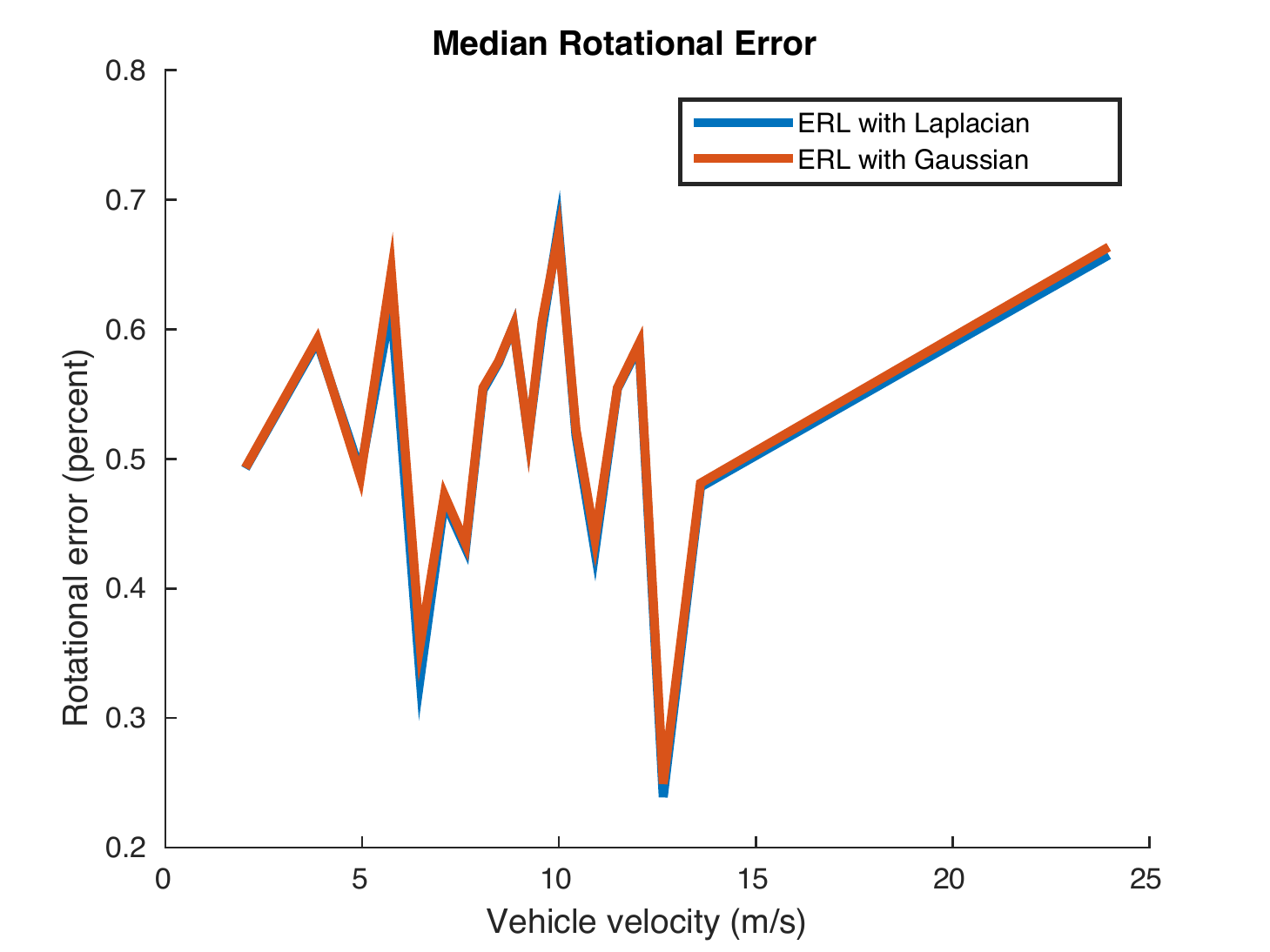}
			}}	
			\caption{Median translational and rotational errors on the full KITTI odometry dataset for ERL with two candidate distributions.}
			\label{fig:likelihoodschemes}
		\end{figure}

		\section{Derivation of Linear Least Squares Estimate}
		
		We first minimize equation (4) from the main paper with respect to the inverse depths $\vect{\rho}$, giving
		\begin{align*}
		&   \min_{\vect{\rho}} E(\vect{t},\vect{\rho},\vect{\omega}) \\
		=&\; \min_{\vect{\rho}} \| A(\vect{t})\vect{\rho} + B\vect{\omega} - u \|_2^2 \\
		=&\; \| -A(\vect{t})\left(A^{\top}(\vect{t}) A(\vect{t})\right)^{-1} A^{\top}(\vect{t})(B\vect{\omega} - u) + B\vect{\omega} - u \|_2^2 \\
		=&\; \| \left(I - A(\vect{t})\left(A^{\top}(\vect{t}) A(\vect{t})\right)^{-1} A^{\top}(\vect{t})\right)(B\vect{\omega} - u) \|_2^2 \\
		=&\; \| A^{\perp}(\vect{t})^{\top} (B\vect{\omega} - u) \|_2^2 .\\
		\end{align*}
		
		We now have an expression in terms of the orthogonal complement of $A(\vect{t})$.
		Finding this orthogonal complement is fairly simple since it is sparse. To show this, we first note that the orthogonal complement of $A(\vect{t})$ is the null space of $A^{\top}(\vect{t})$. $A^{\top}(\vect{t})$ is of the form
		\begin{equation*}
		A^{\top}(\vect{t}) = \begin{bmatrix}
		\vect{t}^{\top} A^{\top}(x_1) &                 0             & \ldots &               0              \\
		0               & \vect{t}^{\top} A^{\top}(x_2) & \ldots &               0              \\
		\vdots             &              \vdots           & \ddots &            \vdots            \\
		0               &                 0             & \ldots & \vect{t}^{\top} A(x_n)^{\top}
		\end{bmatrix} \in \bR^{n \times 2n}.
		\end{equation*}
		Each of the rows of $A^{\top}(\vect{t})$ are orthogonal, so we can consider each of the rows individually. Consider the vector
		\[
		\phi_i = \left( 0,\; 0,\; \ldots,\; 0,\; \vect{t}^{\top} A^{\top}(x_i)J^{\top},\; 0,\; \ldots,\; 0 \right)^{\top}, 
		\]
		where
		\[
		J = \begin{bmatrix} 0 & -1 \\ 1 & 0 \end{bmatrix}
		\]
		is a skew-symmetric matrix in $\bR^{2 \times 2}$. By construction, this vector is orthogonal to the $i^{th}$ column of $A^{\top}(\vect{t})$.  We normalize and concatenate these vectors to form the matrix
		\begin{equation*}
		A^{\perp}(\vect{t}) = \left[ \frac{\phi_1}{\|\phi_1\|} \; \frac{\phi_2}{\|\phi_2\|} \; \ldots \; \frac{\phi_n}{\|\phi_n\|} \right].
		\end{equation*}
		This matrix is very sparse, so we can compute products with it very efficiently: $A^{\perp}(\vect{t})^{\top} B$ and $A^{\perp}(\vect{t})^{\top} u$ can be computed in $\mathcal{O}(n)$ time. From here, the least squares estimate of $\vect{\omega}$ can be computed as:
		\begin{equation}
		\hat{\vect{\omega}}(\vect{t}) = \left( B^{\top} A^{\perp}(\vect{t}) A^{\perp}(\vect{t})^{\top} B \right)^{-1} B^{\top} A^{\perp}(\vect{t}) A^{\perp}(\vect{t})^{\top} u.
		\end{equation}
		In summation notation:
		\begin{equation}
		\begin{split}
		\hat{\vect{\omega}}(\vect{t}) =&\;
		\left( \sum_{i=1}^n \frac{B^{\top}(x_i) JA(x_i) \vect{t} \vect{t}^{\top} A^{\top}(x_i)J^{\top} B(x_i)}{\|JA(x_i) V\|^2} \right)^{-1} \\
		&\;\qquad\left(\sum_{i=1}^n \frac{B^{\top}(x_i) JA(x_i) \vect{t} \vect{t}^{\top} A^{\top}(x_i)J^{\top} u_i}{\|JA(x_i) \vect{t}\|^2} \right).
		\end{split}
		\end{equation}
		There are $2n$ terms to compute, and one inversion of a 3 by 3 matrix, making this $\mathcal{O}(n)$ time to compute. Taken altogether, we compute the residual given $\vect{t}$ by
		\begin{equation*}
		\| A^{\perp}(\vect{t})^{\top} (B\hat{\vect{\omega}}(\vect{t}) - u) \|_2^2.
		\end{equation*}
		
		To compute the residual, we use the error vector $E$, defined as:
		\begin{align}
		E_i(\vect{t}) =&\; \frac{\vect{t}^{\top} A^{\top}(x_i)J^{\top} }{\|JA(x_i) \vect{t}\|} \left(B(x_i)\hat{\vect{\omega}}(\vect{t}) - u_i \right) & \text{for } i = 1, \ldots, n \\
		\label{error_vec_eqn}
		E(\vect{t}) =&\; (E_1(\vect{t}), E_2(\vect{t}), \ldots, E_n(\vect{t}))^{\top} .
		\end{align}
		
		The residual is exactly $||E(\vect{t})||^2 = \sum_i ||E_i(\vect{t})||^2$. As there are $n$ of these error terms, and $\hat{\vect{\omega}}$ takes $\mathcal{O}(n)$ to compute, this residual calculation takes $\mathcal{O}(n)$ to compute. This was shown to be an unbiased estimator in \cite{zhang_fast_1999}.
		
		\section{Lifted Weights Formulation}
		Now, since the cost function is given as a sum of squares, we can optimize it using a Gauss-Newton framework. Therefore, to reject outliers we can use \cite{zach_robust_2014} to optimize this efficiently. Fixing the $\vect{t}$ term, the equation becomes linear:
		\[ \min_{\vect{\omega}} \| A^{\perp}(\vect{t})^{\top} B \vect{\omega} - A^{\perp} u \|_2^2 = \| \vect{f}(\vect{\omega}) \|_2^2 .\]
		So we see the Jacobian is given by $\nabla \vect{f}(\vect{\omega}) = A^{\perp}(\vect{t})^{\top} B$ and thus, as in \cite{zach_robust_2014}, our lifted cost function takes the form
		\[
		\min_{\vect{\omega},\vect{w}} \left\| \begin{pmatrix}
		\vect{w} \circ \vect{f}(\vect{\omega}) \\ \kappa(\vect{w} \circ \vect{w})
		\end{pmatrix} \right\|_2^2.
		\]
		Here we use the smooth truncated quadratic for our $\kappa$ function (applied elementwise):
		\[ \kappa(w^2) = \frac{\tau}{\sqrt{2}} (w^2 - 1), \]
		where $\tau$ is a hyperparameter. Therefore the Jacobian used for the Gauss-Newton iteration is:
		\[
		\hat{\mathbf{J}} = \begin{pmatrix}
		\mathrm{diag}(\vect{w}) \nabla \vect{f}(\vect{\omega}) & \mathrm{diag}(\vect{f}(\omega)) \\
		\mathbf{0} & \nabla \kappa(\vect{w}\circ\vect{w}),
		\end{pmatrix}
		\]
		where $\nabla f$ and $\nabla \kappa$ denote the Jacobian of $f$ and $\kappa$, respectively. From here, we follow the derivation in \cite{zach_robust_2014}.

		\section{Implementation Details of Soatto/Brockett Algorithm}
		\subsection{Expression of $\hat{\vect{\omega}}$}
		Recall equation (2). We rewrite this as:
		\begin{align}
		\hat{\vect{\omega}}(\vect{t}) =&\; G_{full}(\vect{t})^{-1} H_{full}(\vect{t}) \\
		G_{full}(\vect{t}) =&\; \sum_{i=1}^n \frac{B^{\top}(x_i) JA(x_i) \vect{t} \vect{t}^{\top} A^{\top}(x_i)J^{\top} B(x_i)}{\|JA(x_i) V\|^2} \\
		H_{full}(\vect{t}) =&\; \sum_{i=1}^n \frac{B^{\top}(x_i) JA(x_i) \vect{t} \vect{t}^{\top} A^{\top}(x_i)J^{\top} u_i}{\|JA(x_i) \vect{t}\|^2}.
		\end{align}
		As in \cite{chiuso_optimal_2000}, we drop the denominator terms. This gives us:
		The first term we need to consider is the 3 by 3 matrix we need to invert. 
		\begin{align}
		\hat{\vect{\omega}}(\vect{t}) =&\; G(\vect{t})^{-1} H(\vect{t}) \\
		G(\vect{t}) =&\; \sum_{i=1}^n B^{\top}(x_i) JA(x_i) \vect{t} \vect{t}^{\top} A^{\top}(x_i)J^{\top} B(x_i) \\
		H(\vect{t}) =&\; \sum_{i=1}^n B^{\top}(x_i) JA(x_i) \vect{t} \vect{t}^{\top} A^{\top}(x_i)J^{\top} u_i.
		\end{align}
		We focus on $G(\vect{t})$ first. We can write this out in terms of quadratic terms of $\vect{t}\vect{t}^\top$ by introducing the matrices $S^{ij}$, defined as:
		\begin{equation*}
		S^{ij}_{kl} = \begin{cases}
		1 & \textrm{ if } i = k, j = l, \textrm{ or } i = l, j = k \\
		0 & \textrm{ otherwise }
		\end{cases}
		\end{equation*}
		\begin{align*}
		G(\vect{t})
		=&\;\; \vect{t}_1^2
		\left( \sum_{i=1}^n B^{\top}(x_i) JA(x_i) S^{11} A^{\top}(x_i)J^{\top} B(x_i) \right) \\
		&+\; \vect{t}_1 \vect{t}_2
		\left( \sum_{i=1}^n B^{\top}(x_i) JA(x_i) S^{12} A^{\top}(x_i)J^{\top} B(x_i) \right) \\
		&+\; \vect{t}_1 \vect{t}_3
		\left( \sum_{i=1}^n B^{\top}(x_i) JA(x_i) S^{13} A^{\top}(x_i)J^{\top} B(x_i) \right) \\
		&+\; \vect{t}_2^2
		\left( \sum_{i=1}^n B^{\top}(x_i) JA(x_i) S^{22} A^{\top}(x_i)J^{\top} B(x_i) \right) \\
		&+\; \vect{t}_2 \vect{t}_3
		\left( \sum_{i=1}^n B^{\top}(x_i) JA(x_i) S^{23} A^{\top}(x_i)J^{\top} B(x_i) \right) \\
		&+\; \vect{t}_3^2
		\left( \sum_{i=1}^n B^{\top}(x_i) JA(x_i) S^{33} A^{\top}(x_i)J^{\top} B(x_i) \right) \\
		=&\; \sum_{i < j} \vect{t}_i \vect{t}_j G^{ij}, \\
		\end{align*}
		where $G^{ij}$ is defined appropriately.
		We will use $G^{ij}_k$ to denote the $k^{th}$ column of $G^{ij}$. We know that the inverse of a 3 by 3 matrix with columns $c_1$, $c_2$, $c_3$ has an inverse given by
		\begin{equation*}
		\frac{1}{\det([c_1 \; c_2 \; c_3])}
		\begin{bmatrix} (c_2 \times c_3)^{\top} \\ (c_3 \times c_1)^{\top} \\ (c_1 \times c_2)^{\top} \end{bmatrix}.
		\end{equation*} 
		We also know that the cross product is bi-linear, so from this we can write out the inverse of $G$ analytically.
		\begin{align*}
		& G^{-1}(\vect{t}) \\
		=&\; \frac{1}{\det(G(\vect{t}))} \begin{bmatrix}
		\left(\left(\sum_{i < j} \vect{t}_i \vect{t}_j G^{ij}_2 \right) \times \left(\sum_{k < l} \vect{t}_k \vect{t}_l G^{kl}_3 \right) \right)^{\top} \\ 
		\left(\left(\sum_{i < j} \vect{t}_i \vect{t}_j G^{ij}_3 \right) \times \left(\sum_{k < l} \vect{t}_k \vect{t}_l G^{kl}_1 \right) \right)^{\top} \\ 
		\left(\left(\sum_{i < j} \vect{t}_i \vect{t}_j G^{ij}_1 \right) \times \left(\sum_{k < l} \vect{t}_k \vect{t}_l G^{kl}_2 \right) \right)^{\top} \\ 
		\end{bmatrix} \\
		=&\; \frac{1}{\det(G(\vect{t}))} \begin{bmatrix}
		\sum_{i < j, k < l} \vect{t}_i \vect{t}_j \vect{t}_k \vect{t}_l \left( G^{ij}_2 \times G^{kl}_3 \right)^{\top} \\ 
		\sum_{i < j, k < l} \vect{t}_i \vect{t}_j \vect{t}_k \vect{t}_l \left( G^{ij}_3 \times G^{kl}_1 \right)^{\top} \\ 
		\sum_{i < j, k < l} \vect{t}_i \vect{t}_j \vect{t}_k \vect{t}_l \left( G^{ij}_1 \times G^{kl}_2 \right)^{\top} \\ 
		\end{bmatrix} \\
		=&\; \frac{1}{\det(G(\vect{t}))} \sum_{i < j, k < l} \vect{t}_i \vect{t}_j \vect{t}_k \vect{t}_l \begin{bmatrix}
		\left( G^{ij}_2 \times G^{kl}_3 \right)^{\top} \\ 
		\left( G^{ij}_3 \times G^{kl}_1 \right)^{\top} \\ 
		\left( G^{ij}_1 \times G^{kl}_2 \right)^{\top} \\ 
		\end{bmatrix} .\\
		\end{align*}
		
		The terms in the matrix component become a $4^{th}$ degree polynomial of 3 variables with 15 terms (after grouping) with matrix coefficients. We can also can compute the determinant explicitly using the fact that the determinant of a 3 by 3 matrix is the triple product of its columns. 
		\begin{align*}
		\det(G(\vect{t}))
		=&\; (G_2(\vect{t}) \times G_3(\vect{t}))^{\top} G_1(\vect{t}) \\
		=&\; \left(\sum_{i < j, k < l} \vect{t}_i \vect{t}_j \vect{t}_k \vect{t}_l \left( G^{ij}_2 \times G^{kl}_3 \right)\right)^{\top} \left(\sum_{p < q} \vect{t}_p \vect{t}_q G^{pq}_1 \right) \\
		=&\; \sum_{i < j, k < l, p < q} \vect{t}_i \vect{t}_j \vect{t}_k \vect{t}_l \vect{t}_p \vect{t}_q \left( \left( G^{ij}_2 \times G^{kl}_3 \right)^{\top} G^{pq}_1 \right) \\
		\end{align*}
		After grouping terms, this becomes a $6^{th}$ degree polynomial with 28 terms. This makes each element of $G^{-1}$ a $6^{th}$ degree rational function. \\
		
		\noindent In a similar fashion, we find the expression
		\begin{align*}
		H(\vect{t})
		=&\;\; \vect{t}_1^2
		\left( \sum_{i=1}^n B^{\top}(x_i) JA(x_i) S^{11} A^{\top}(x_i)J^{\top} u_i \right) \\
		&+\; \vect{t}_1 \vect{t}_2
		\left( \sum_{i=1}^n B^{\top}(x_i) JA(x_i) S^{12} A^{\top}(x_i)J^{\top} u_i \right) \\
		&+\; \vect{t}_1 \vect{t}_3
		\left( \sum_{i=1}^n B^{\top}(x_i) JA(x_i) S^{13} A^{\top}(x_i)J^{\top} u_i \right) \\
		&+\; \vect{t}_2^2
		\left( \sum_{i=1}^n B^{\top}(x_i) JA(x_i) S^{22} A^{\top}(x_i)J^{\top} u_i \right) \\
		&+\; \vect{t}_2 \vect{t}_3
		\left( \sum_{i=1}^n B^{\top}(x_i) JA(x_i) S^{23} A^{\top}(x_i)J^{\top} u_i \right) \\
		&+\; \vect{t}_3^2
		\left( \sum_{i=1}^n B^{\top}(x_i) JA(x_i) S^{33} A^{\top}(x_i)J^{\top} u_i \right) \\
		=&\; \sum_{i < j} \vect{t}_i \vect{t}_j H^{ij} \\
		\end{align*}
		This gives the final form of the equation:
		\begin{align}
		\hat{\vect{\omega}}(\vect{t})
		=&\;  \frac{1}{\det(G(\vect{t}))} \sum_{i < j, k < l, p < q} \vect{t}_i \vect{t}_j \vect{t}_k \vect{t}_l \vect{t}_p \vect{t}_q \begin{bmatrix}
		\left( G^{ij}_2 \times G^{kl}_3 \right)^{\top} \\ 
		\left( G^{ij}_3 \times G^{kl}_1 \right)^{\top} \\ 
		\left( G^{ij}_1 \times G^{kl}_2 \right)^{\top} \\ 
		\end{bmatrix} H^{pq}
		\end{align}
		
		\noindent We are left with $\hat{\vect{\omega}}(\vect{t})$ as a degree-6 rational function of $V$, meaning it has 28 terms in the numerator and denominator for each element.
		\subsection{Expression of Cost Function}
		First, we express the cost function as:
		\begin{equation}
		f(\vect{t}) = \sum_i \left( \vect{t}^{\top} A^{\top}(x_i)J^{\top} \left(B(x_i)\hat{\vect{\omega}}(\vect{t}) - u_i \right) \right)^2
		\end{equation}
		Now we expand and simplify this by plugging in the definitions given above for $G$ and $H$:
		\begin{align*}
		f(\vect{t})
		=&\; \sum_i \left( \vect{t}^{\top} A^{\top}(x_i)J^{\top} \left(B(x_i)\hat{\vect{\omega}}(\vect{t}) - u_i \right) \right)^2 \\
		=&\; \sum_i \left(B(x_i)\hat{\vect{\omega}}(\vect{t}) - u_i\right)^\top JA(x_i) \vect{t}\vect{t}^{\top} A^{\top}(x_i)J^{\top} \left(B(x_i)\hat{\vect{\omega}}(\vect{t}) - u_i\right) \\
		=&\; \hat{\vect{\omega}}(\vect{t})^{\top} \left( \sum_i B^\top(x_i)JA(x_i) \vect{t}\vect{t}^{\top} A^{\top}(x_i)J^{\top} B(x_i) \right) \hat{\vect{\omega}}(\vect{t}) \\
		&\;\qquad - \left( \sum_i  B^{\top}(x_i) JA(x_i) \vect{t}\vect{t}^{\top} A^{\top}(x_i)J^{\top} u_i \right)^\top \hat{\vect{\omega}}(\vect{t}) \\
		&\;\qquad + \vect{t}^{\top} \left( \sum_i  A^{\top}(x_i)J^{\top} u_i u_i^\top JA(x_i) \right) T \\
		=&\; \hat{\vect{\omega}}(\vect{t})^{\top} G(\vect{t}) \hat{\vect{\omega}}(\vect{t}) - 2 H(\vect{t})^{\top} \hat{\vect{\omega}}(\vect{t}) + \vect{t}^{\top} S \vect{t}  \\
		=&\; \hat{\vect{\omega}}(\vect{t})^{\top} G(\vect{t}) (G^{-1}(\vect{t}) H(\vect{t})) - 2 H(\vect{t})^{\top} \hat{\vect{\omega}}(\vect{t}) + \vect{t}^{\top} S \vect{t}  \\
		=&\; \hat{\vect{\omega}}(\vect{t})^{\top} H(\vect{t}) - 2 H(\vect{t})^{\top} \hat{\vect{\omega}}(\vect{t}) + \vect{t}^{\top} S \vect{t}  \\
		=&\; \vect{t}^{\top} S \vect{t} - H(\vect{t})^{\top} \hat{\vect{\omega}}(\vect{t}), \\
		\end{align*}
		where
		\[ S = \sum_i  A^{\top}(x_i)J^{\top} u_i u_i^\top JA(x_i). \]
		This gives us the final equation
		\begin{equation}
		f(\vect{t}) = \vect{t}^{\top} S \vect{t} - H(\vect{t})^{\top} \hat{\vect{\omega}}(\vect{t}). \\
		\end{equation}

\end{document}